# An Empirical Evaluation of Large Language Models on Consumer Health Questions

Moaiz Abrar[1,2], Yusuf Sermet[1], Ibrahim Demir[3,4]

[1] IIHR Hydroscience and Engineering, University of Iowa
[2] Computer Science, University of Iowa
[3] River-Coastal Science and Engineering, Tulane University
[4] ByWater Institute, Tulane University

**Abstract**
This study evaluates the performance of several Large Language Models (LLMs) on MedRedQA, a dataset of consumer-based medical questions and answers by verified experts extracted from the AskDocs subreddit. While LLMs have shown proficiency in clinical question answering (QA) benchmarks, their effectiveness on real-world, consumer-based, medical questions remains less understood. MedRedQA presents unique challenges, such as informal language and the need for precise responses suited to non-specialist queries. To assess model performance, responses were generated using five LLMs: GPT-4o mini, Llama 3.1: 70B, Mistral-123B, Mistral-7B, and Gemini-Flash. A cross-evaluation method was used, where each model evaluated its responses as well as those of others to minimize bias. The results indicated that GPT-4o mini achieved the highest alignment with expert responses according to four out of the five models' judges, while Mistral-7B scored lowest according to three out of five models' judges. This study highlights the potential and limitations of current LLMs for consumer health medical question answering, indicating avenues for further development.

# 1. Introduction

In healthcare consultations, clinicians raise questions related to patient care but only find answers to half of those questions due to limited time or the belief that an answer may not exist (Del Fiol et al., 2014). Medical question-answering (QA) systems have the potential to address these problems by giving fast responses to clinicians' questions. These systems are designed to provide accurate and relevant answers to medical queries by leveraging natural language processing techniques. Traditional systems in this domain typically utilize information retrieval techniques to draw responses from structured medical databases or relevant documents. These systems often involve classifying question type, such as Yes/No or factual questions (Sarrouti et al., 2015), and then employ semantic matching and extraction methods to generate concise responses from the matched documents.

Incorporating artificial intelligence (AI) and knowledge graphs, including ontologies (Baydaroglu et al., 2023), into educational and healthcare domains provides innovative avenues to enhance communication and knowledge access. AI-driven systems use knowledge graphs to organize and interlink vast arrays of information, enabling both educators (Sajja et al., 2024a; 2024b) and healthcare providers to access tailored, contextually relevant data (Chi et al., 2020). When integrated with chatbots, these systems can facilitate interactive and personalized learning experiences in education, offering students immediate, accurate responses to their inquiries. In healthcare, chatbots powered by AI and ontologies can assist in patient triage, symptom checking (Chi et al., 2023), and patient education, ensuring that users receive up-to-date medical information efficiently (Sermet and Demir, 2021). By leveraging these technologies, educational and healthcare chatbots can move beyond simple transactional interactions to deliver sophisticated, nuanced assistance that supports both learning and clinical decision-making processes (Pursnani et al., 2023).

The advent of Large Language Models (LLMs) such as GPT-4 (OpenAI, 2023) has introduced transformative possibilities for many use cases in education (Sajja et al., 2023a; 2023b), operational support (Samuel et al., 2024), and health care. These models, for example, can address the significant documentation burden in EHRs by automating text summarization, allowing clinicians to review condensed, relevant summaries rather than lengthy clinical notes (Jain et al., 2024). Several studies have also been conducted to explore the potential of AI and LLMs in medical support and QA (Zhang et al., 2023, Banerjee et al., 2023; Wang et al., 2024).

Current LLMs have been extensively evaluated on clinical QA benchmarks, such as MedQA (Chai et al., 2020) and PubMedQA (Jin et al., 2019), where questions are typically structured in a multiple-choice format to assess clinical accuracy and factual recall. However, multiple-choice QA does not fully capture the complexity of real-world medical inquiries, as it limits responses to predefined options and restricts the model's ability to provide nuanced, explanatory answers. To address these limitations, these datasets have transformed into open-ended question formats, allowing models to handle more elaborate responses that better reflect the complexities of clinical scenarios (Yang et al., 2024; Gopal et al., 2024). Additionally, LLMs have been evaluated on qualities beyond factual accuracy, such as safety, bias, and language understanding

(Kanithi et al., 2024), to better align with the complexities encountered in real-world medical interactions.

Open-ended clinical QA benchmarks, however, are focused on structured, professional queries rather than consumer-based questions. Consumer queries typically lack specific medical terminology (Welivita & Pu, 2023), use informal language, and may pose open-ended inquiries with limited or ambiguous detail. This difference from clinical-style questions presents a unique challenge for LLMs, which must interpret and respond to questions in a way that accommodates the informal, varied nature of consumer inquiries. Previous work in the consumer medical QA domain includes the CHiQA system (Demner-Fushman et al., 2020), which focuses on reliable information retrieval from consumer-friendly sources such as MedlinePlus (MedlinePlus, n.d.) to address common health questions. By using trusted patient-oriented sources, this system bridges the gap between consumer queries and trustworthy medical content, though it still encounters challenges in matching informal consumer language with precise medical information.

Additionally, research on improving consumer medical QA demonstrates the difficulty consumers face when formulating specific questions that align with their informational needs (Nguyen, 2024). Work by Nguyen addresses this issue by proposing improved biomedical representational learning and statistical keyword modeling. These improvements aid in retrieving medical answers even when consumer questions are vague or contain informal language (Nguyen, 2024). Another study conducted experiments with a QA system to retrieve answers for real consumer medication queries (Abacha et al., 2019). By curating a dataset containing genuine consumer questions about medications with corresponding expert verified answers, researchers observed that the QA system struggled with retrieving the correct answers. They also highlighted the need for better contextual understanding in consumer medication QA.

Several existing studies have also assessed LLM responses to questions posted on the AskDocs (Reddit, 2024) subreddit. For example, in (Ayers et al., 2023) the authors compared physician and ChatGPT (ChatGPT, n.d.) responses to 195 randomly selected questions from the subreddit, finding that healthcare professionals preferred ChatGPT's responses in 78.6% of 585 evaluations. Other studies have examined LLM responses in specialized fields, such as Otolaryngology (Carnino et al., 2024), where ChatGPT's responses to 15 domain-specific questions were rated with an accuracy of 3.76 out of 5, and Cancer (Chen et al., 2024), where physicians evaluated 200 cancer-related questions and rated LLM responses to be of higher quality.

While these studies are valuable in understanding LLM capabilities in answering consumer health queries, there remains a lack of evaluation of LLMs on a large-scale dataset of consumer health questions. The MedRedQA (Nguyen et al., 2023) dataset addresses this gap by providing a large collection of consumer-based medical questions and expert answers extracted from the AskDocs subreddit. This dataset includes a wide range of layperson queries on medical topics, offering an opportunity to evaluate LLMs on a large number of real-world, consumer-oriented healthcare questions in non-clinical settings. Physician answers in the dataset (extracted from

AskDocs) are used as ground truths in this study. This is justified by the verification of the physicians' credentials by the subreddit's moderators.

This study aims to evaluate the effectiveness of five prominent LLMs in answering consumer-based medical questions from the MedRedQA dataset, which includes real-world queries and expert responses extracted from the AskDocs subreddit. The LLMs assessed are GPT-4o mini (OpenAI, 2024), Llama 3.1 (70B) (Llama Team AI @ Meta, 2024), Mistral-123B (Mistral AI, 2024), Mistral-7B (Jiang et al., 2023), and Gemini-Flash (Georgiev et al., 2024). Employing a cross-evaluation approach, each model's responses were not only evaluated by itself but also by the other models to minimize bias in the assessment process. The findings reveal that GPT-4o mini achieved the highest alignment with expert answers according to four of the five model judges, whereas Mistral-7B scored the lowest according to three of the five judges. These results highlight both the potential and current limitations of LLMs in addressing consumer health questions, indicating important avenues for future development.

The rest of the article is organized as follows. The Methods section describes the selection of models, the rationale for choosing the MedRedQA dataset, the prompt generation strategies, and the evaluation techniques utilized in the study. The Results and Discussion section presents the evaluation outcomes, analyzes the performance of each model, and discusses the implications of the cross-evaluation findings, including considerations about model sizes and evaluator reliability. The Limitations section acknowledges the constraints faced during the study, such as dataset challenges and evaluation complexities. Finally, the Conclusions and Future Work section summarizes the key insights and outlines potential directions for enhancing LLM performance in consumer health question answering.

## 2. Methods

The focus of this study is the evaluation of LLM responses to consumer-based medical questions. The following sections describe the scope and methods used in this study.

### 2.1. Purpose and Scope

This study addresses the gap in understanding the effectiveness of LLMs in answering consumer-based medical questions. Existing research has predominantly focused on multiple-choice medical question answering datasets or smaller, sample-based studies on questions extracted from AskDocs. This study evaluates LLMs on MedRedQA—a dataset consisting of a large number of real-world, consumer-oriented medical inquiries extracted from AskDocs. To guide this exploration, two primary research questions are formulated:

*RQ1:* How effectively do current LLMs perform in answering consumer-based medical questions in MedRedQA?

*RQ2:* How reliable are different LLMs as evaluators of their own and other models' responses to medical questions, and does evaluator reliability vary significantly across models?

*RQ1* seeks to determine the extent to which LLMs can generate responses aligned with expert answers in the dataset. *RQ2* seeks to determine whether LLMs can act as reliable evaluators and whether the choice of evaluator can affect the evaluation results. Given the computational cost and resource constraints associated with using larger models, this study focuses on smaller, more cost-effective models. This practical focus ensures that the findings of the study remain accessible and relevant to future researchers exploring cost-effective methods for medical question answering research.

## 2.2. Model Selection

The following five LLMs were selected for this study based on their demonstrated effectiveness in natural language processing tasks. These models provide a diverse selection from both open-source and proprietary sources, capturing a broad view of current LLM capabilities for consumer medical QA.

- *GPT-4o mini*: A compact variant of the GPT-4o model that prioritizes efficiency while maintaining strong reasoning capabilities.

- *Llama 3.1: 70B*: Part of the Llama model family, this 70-billion parameter model performs competitively with other compact models and demonstrates robust language understanding across a range of benchmarks.

- *Gemini-Flash*: This model is part of the Gemini 1.5 series and is developed as a more cost-effective and faster alternative to Gemini-1.5 Pro (Georgiev et al., 2024).

- *Mistral 7B and Mistral 123B*: Known for outperforming larger models on several benchmarks, the Mistral family offers powerful small-scale models that demonstrate competitive performance for their size.

## 2.3. Dataset Selection

This section describes some of the datasets commonly used in medical question answering research and why the MedRedQA dataset is selected for this study over other datasets.

### 2.3.1. Existing Datasets in Medical QA

Several datasets have been used traditionally to evaluate the capabilities of large language models in medical question-answering tasks. In multiple choice question-answering, the most prominent ones include MedQA, PubMedQA, and MMLU-Clinical Knowledge (Hendrycks et al., 2020).

*MedQA*: This dataset is collected from US medical licensing exams (USMLE) and includes questions designed to test structured medical knowledge. The dataset is often used for evaluating the clinical knowledge of LLMs, with models such as GPT-4 and Med-Gemini (Wang et al., 2024) achieving accuracies above 90% (Banerjee et al., 2023; Wang et al., 2024). It uses a multiple-choice format, which is useful for testing medical knowledge but is less relevant for real-world, open-ended medical queries.

*PubMedQA*: It consists of clinical research questions extracted from the PubMed database, where answers can take the form of abstracts, yes/no responses, or specific medical conclusions. While the benchmark allows for both short-form and long-form responses, it remains structured around formal medical research rather than the informal queries typical of consumer healthcare.

*MMLU-Clinical Knowledge*: This benchmark tests a model's knowledge across multiple domains, including medicine, using a similar multiple-choice format. Like MedQA, this dataset focuses on assessing clinical and factual knowledge.

To address the limitations of multiple-choice formats, some datasets have been transformed or created to require open-ended, detailed answers. These include MedQA-Open (Gopal et al., 2024), MedQA-CS (Yao et al., 2024), and MedQuAD (Abacha & Demner-Fushman, 2019):

*MedQA-Open*: It is a modified version of MedQA, adapted to require models to generate open-ended answers. While it allows for detailed responses, the questions remain based on medical licensing exams, limiting their relevance to consumer-based, informal queries.

*MedQA-CS*: This benchmark focuses on clinical skills, modeled after the medical education's Objective Structured Clinical Examinations (OSCEs). This dataset evaluates LLMs through two tasks: LLM-as-medical-student and LLM-as-clinical-examiner, both reflecting formal clinical scenarios. It provides an assessment of LLMs in settings that are closer to real-world clinical scenarios. However, the benchmark is less relevant for consumer-based medical question-answering tasks due to its focus on professional clinical settings.

*MedQuAD*: It contains question-answer pairs extracted from National Institutes of Health (NIH) websites. The dataset includes detailed, structured answers based on expert medical content. However, like other datasets, it does not reflect the informal nature of consumer healthcare questions.

These datasets are crucial for evaluating how well LLMs can handle open-ended questions in clinical scenarios. However, their reliance on formal medical cases or clinical exam formats makes them less suitable for assessing how models respond to consumer-facing queries, which often lack medical precision or structure.

### 2.3.2. MedRedQA

This dataset includes 51,000 pairs of consumer medical questions and expert answers extracted from the AskDocs subreddit. AskDocs allows consumers to post health-related questions, and only verified medical professionals provide answers. The dataset has two parts, one consisting of samples where expert responses include citations to PubMed articles and the other without any citations. The second part of the dataset is used in this study. The test set of this dataset contains 5099 samples. Each sample contains a title, the body, the response by the medical expert, the response score, and the occupation of the expert. The dataset includes responses that received the highest upvotes (response score), reflecting a consensus on the relevance and quality of the answers.

MedRedQA is used for this study because it provides a large set of real consumer healthcare queries which LLMs can be evaluated on. The informal nature of these questions presents a unique challenge for LLMs and provides a benchmark to evaluate the ability of models to provide accurate answers to medical queries in non-clinical settings.

**2.4. Prompt Generation**

Two distinct prompts were used in this study, one for the response to user questions, and one for the evaluation of LLM responses for agreement with physician responses. The two prompts are shown in Table 1.

Table 1. Prompts Used for generating answers (RQ1) and evaluating answers (RQ2)

| **RQ1 Prompt** | **RQ2 Prompt** |
| --- | --- |
| <u>System Instructions:</u> You are able to understand medical questions and provide precise answers to them.<br><br><u>Prompt:</u> Then different user prompts are tested to make sure that the model responds as required, with answers that are precise and that don't have additional commentary.<br><br>The following 'user' prompt is given to each of the five models:<br><br>"I will provide you with a medical question and the title associated with it. You will answer that question as precisely as possible, addressing only what is asked. There is no need to provide additional context and details." | <u>System Instructions:</u> You are able to understand medical content and answer any queries regarding the content.<br><br><u>Prompt:</u> I will provide you with a medical question, its associated title, and two responses to that question: one from a medical expert (which is the correct answer or ground truth) and another from a different source. Your task is to compare the information in the other response with the expert's, treating the expert's answer as the ground truth.<br><br>You are not evaluating the correctness of the other response directly. Instead, your focus is solely on how closely the information in the other response aligns with the information in the expert's response (which is the correct answer).<br><br>Your output must strictly be one of the following two words:<br>1. 'Agree' if the main information in both responses is the same.<br>2. 'Disagree' if the main information in the other response is not similar to the expert's.<br><br>Your output should consist of only one of these two terms, without any additional text |

For both cases, different prompts are tested to make sure that the model responds as required, with answers that are precise and that don't have additional commentary. For the RQ1 prompt, after getting the acknowledgement from the model, the title and body of the question are provided to the model to generate the answer. The title is included because it can often contain information that is important to answer the question. For the RQ2 prompt, after getting the acknowledgement from the model, the title, body, and the expert and model answers are provided. The acknowledgement message from the models is obtained only once and the same one is used when evaluating each of the samples.

The Mistral-7B model is provided with the same acknowledgement message as the Mistral-123B model, instead of its own, because the model struggles with understanding the instructions properly. Specifically, the model started to create its own example and respond with "Agree" or "Disagree". The acknowledgement from the Mistral-123B model is used to make sure that the model responds correctly when provided with the two responses to compare. Despite explicit instructions in the prompt that the output should be "Agree" or "Disagree", responses from the Mistral models are of the form " Agree" or " Disagree" and responses from the Gemini-Flash model are of the form "Agree \n" or "Disagree \n". These responses are treated as "Agree" or "Disagree" respectively. Responses other than these are classified as "other".

### 2.5. Evaluation of Responses

Responses generated by LLMs are evaluated through a cross-model approach, where each model evaluates its own output and is cross-evaluated by all other models to reduce bias. In the evaluation process models are instructed to classify responses as either "Agree" or "Disagree" based on their similarity to expert-provided answers. This approach was chosen over traditional metrics such as ROUGE (Lin, 2004) and BERT-SCORE (Zhang et al., 2019) because these metrics primarily measure surface-level lexical similarities, which may not reflect deeper semantic alignment. Even when the wording between two answers differs, the core information can still be highly aligned, which LLM-based evaluation may be able to better capture.

LLMs have been successfully used as evaluators in multiple research contexts, including pairwise comparisons, where they assess responses based on factors such as helpfulness, fluency, and factual accuracy (Zhang et al., 2024; Levy et al., 2024). Other studies have demonstrated the effectiveness of LLMs as evaluators in comparing expert responses with LLM generated responses in the medical domain (Chen et al., 2024). These prior successes make LLMs suitable evaluators for this study.

LLM generated answers are categorized as either "Agree" or "Disagree". While some evaluation methodologies include a "Neutral" category to account for responses that are neither fully correct nor incorrect, the category isn't included in this evaluation because preliminary experiments showed that most of the answers were classified as "Neutral". This could be because model responses do not perfectly match expert answers, which makes the models classify most samples as Neutrals. Using a binary evaluation instead forces the models to offer clearer distinctions, improving the utility of the results. Furthermore, this evaluation assumes the accuracy of expert answers as only verified individuals can respond to questions.

## 3. Results and Discussion

Evaluation results are shown in Figure 1. GPT-4o mini responses achieved the highest percentage agreement with expert answers as evaluated by four out of the five model judges. Results also show that Mistral-7B tends to give higher agreement scores as an evaluator across the board, possibly indicating a bias toward lenient evaluations. On the other hand, the Gemini-Flash and Mistral-123B models tend to provide lower agreement scores, suggesting that these models are more critical evaluators. It should also be noted that the agreement and disagreement scores don't sum up to 100 except when the evaluator is GPT-4o mini. This is because there is a small percentage of responses for which the models either don't provide answers or provide answers other than "Agree" or "Disagree". Since the number of such responses is very small, it doesn't affect the results by a significant amount.

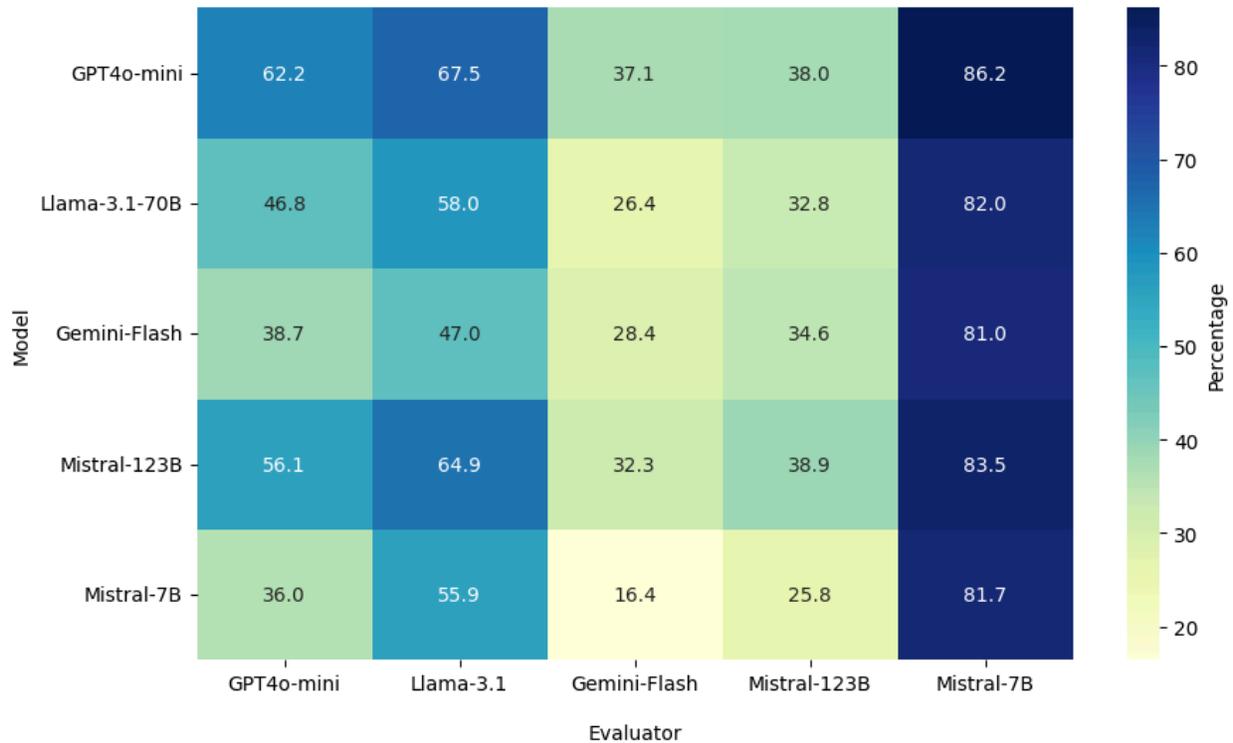

Figure 1. Cross Model Evaluation Results of LLM Responses

To assess (a.k.a, judge) the effectiveness of each model as an evaluator, a sample of 50 expert-model response pairs was selected for manual review, yielding 250 total evaluation pairs across the five models. The sample was curated to include questions with straightforward, unambiguous answers, which allowed accurate assessments without requiring specialized medical knowledge. Complex questions that necessitated in-depth medical expertise for comparison were excluded to maintain the evaluation's reliability.

To minimize bias, the evaluators' initial classifications were not visible during manual evaluation. As shown in Table 2, the Gemini-Flash model achieved the highest accuracy as an evaluator, scoring 77.2%, while Mistral-7B showed the lowest performance at 51.6%. Table 3 shows agreement scores for each model using Gemini-Flash as the evaluator, with GPT-4o mini achieving the highest score (37.1%).

*Safety Mechanisms in Gemini-Flash:* Experiments revealed that the Gemini-Flash model sometimes refused to answer questions entirely due to its built-in safety mechanisms, instead returning a "safety error". Moreover, the model often responded by saying that "it was an AI model and could not offer medical advice" and in one particular example which involved surgery, the model refused to provide an answer saying instead that it was a "dangerous procedure" and asked to seek professional medical help. This resulted in a lot of disagreement outputs, and contributed to the low agreement score for the model as compared to other models.

Table 2: Evaluation scores for models as judges

| Model | Evaluation Score (%) |
|---|---|
| gpt4o-mini | 76.0 |
| llama-3.1-70B | 72.0 |
| gemini-flash | 77.2 |
| mistral-123B | 74.4 |
| mistral-7B | 51.6 |

Table 3: Agreement scores with Gemini-Flash

| Model | Agreement Score (%) |
|---|---|
| gpt4o-mini | 37.1 |
| llama-3.1-70B | 26.4 |
| gemini-flash | 28.4 |
| mistral-123B | 32.3 |
| mistral-7B | 16.4 |

*Discrepancies Across Evaluation Scores:* The results show variations in model behavior as evaluators, where Mistral-7B shows a tendency toward lenient assessments resulting in scores greater than 80% for all models. On the other hand, the Gemini-Flash and Mistral-123B models are more critical, providing accuracy scores lower than 40% across all models. Evaluation results for the models as judges indicate that the Mistral-7B model is the worst performing evaluator, which explains its high agreement scores across all evaluations. Gemini-Flash performs the best as an evaluator (Table 2) but has the second lowest average agreement score (37.2%). This difference suggests that, although Gemini-Flash's safety mechanisms lead to lower agreement scores due to frequent refusals to answer, these same mechanisms do not affect its ability to evaluate the responses of other models.

*Model Sizes:* Mistral-7B achieves the lowest agreement score and the lowest accuracy as an evaluator. This might suggest that closed models (GPT4o-mini, Gemini-Flash) are closer in size to Llama 3.1-70B and Mistral-123B than Mistral-7B, as similarly sized models would be expected to exhibit comparable performance levels. The significantly lower performance of

Mistral-7B may be attributed to its smaller size, which could limit its ability to answer consumer based medical questions accurately. However, whether or not specialized fine-tune models smaller in size exhibit the same patterns is planned for future work.

*Low Average Agreement Scores:* Low average agreement scores, as shown in Table 4, emphasize the difficulties current LLMs face in providing accurate answers to consumer-based medical questions. Excluding the Mistral-7B model due to its relatively poor evaluator performance, we observed that even the highest-performing model, GPT4o-mini, achieved only 51.2% accuracy. This finding highlights the challenges mid or small sized LLMs encounter when attempting to address medical inquiries posed by consumers. Future work should consider fine-tuning LLMs specifically on consumer-based medical question answering datasets. Specialized models fine-tuned on medical data have shown to improve performance on medical datasets, for example Med-Gemini currently achieves the highest accuracy (91.1%) on MedQA (Carnino et al., 2024). Fine-tuning may also improve the accuracy of model responses to consumer medical queries.

Table 4: Average Agreement Scores

| Model | Average Agreement Scores (%) |
|---|---|
| gpt4o-mini | **51.2** |
| llama-3,1-70B | 41.0 |
| gemini-flash | 37.2 |
| mistral-123B | 48.1 |
| mistral-7B | 33.5 |

Another promising approach is retrieval-augmented generation (RAG). RAG combines LLMs with information retrieval techniques to pull relevant data from external medical sources, such as MedlinePlus (MedlinePlus, n.d.), before generating an answer. This approach allows models to use information outside its learned parameters to generate responses to questions. RAG has shown promising results in improving the accuracy of LLMs for medical question answering (Xiong et al., 2024). Consumer-facing QA systems may also benefit by augmenting LLMs with external medical knowledge before responding to queries. The results of this study also show that LLMs should be evaluated on broader datasets in order to better understand the limitations and capabilities of models in this domain in the real-world.

## 3.1. Limitations

This section outlines the limitations of this study, specifically focusing on challenges related to the MedRedQA dataset and the evaluation process used.

*Incomplete Questions:* One limitation of this study lies in the nature of the MedRedQA dataset, which includes instances where expert responses prompt the user for additional information or clarity. These situations often involve requests for supplementary details or, at times, visual inputs, such as clear images, which are not included in the dataset due to privacy restrictions. As a result, model responses are sometimes misaligned with expert answers, especially when interpreting cases where an image would provide critical context.

This limitation also extends to cases where expert answers hinge on situational or time-specific knowledge. For example, questions related to health protocols during the COVID-19 pandemic may lack explicit mention of the pandemic context, yet experts assume this context in their responses. Models, however, may not interpret these questions correctly without this contextual indicator, potentially leading to inaccurate or irrelevant responses. An example is questions such as whether it is safe to bring elderly individuals to hospitals during the COVID period; without a clear indication in the question, models may miss the situational implications present in the expert responses.

*Sample Size for Judging Evaluators:* Another limitation is the small sample size used to manually judge evaluator performance. The limited scope, focused primarily on straightforward questions, may not reflect the models' true evaluator capabilities in more complex, ambiguous cases. Consequently, expanding this sample and incorporating a diverse range of medical question types in future studies would enhance the generalizability and reliability of evaluator judgments. It would also allow us to select models best suited for evaluations and increase confidence in the results of model performance comparisons.

*Credibility of Physician Responses:* The study assumes that physician responses are credible based on the verification done by the subreddit's moderators. However, this verification might not necessarily imply that the answers by the physicians are always correct.

## 4. Conclusions and Future Work

LLMs have enormous potential in transforming the medical question answering domain. In the clinical QA domain, they could help physicians quickly get answers to their queries and consequently improve the quality of care that they offer to their patients. In the consumer QA domain, they could help consumers get the preliminary guidance that they often seek before deciding to go for a hospital visit. However, current LLMs are not yet capable of being deployed for real-world consumer medical questions, answering applications in a zero-shot manner.

As the results of this study show, LLMs struggle in providing accurate answers to consumer health questions. There is a need to extensively evaluate LLMs on a broader range of consumer QA benchmarks that reflect real-world healthcare scenarios in order to accurately assess their reliability in providing answers to consumer questions. The findings of this study highlight both the potential and the current limitations of using LLMs for consumer health question answering. While models like GPT-4o mini show promise, the overall low agreement scores indicate that significant improvements are needed before LLMs can be reliably deployed in real-world healthcare settings.

One immediate avenue for improvement is the fine-tuning of LLMs on datasets specifically curated for consumer health questions. As shown in other domains, specialized fine-tuning can substantially enhance a model's performance by allowing it to better understand the nuances and linguistic patterns typical of the target domain. Developing models that are trained on large-scale, diverse datasets like MedRedQA could help bridge the gap between current capabilities and the requirements for effective consumer health assistance. Furthermore, expanding LLMs to handle multimodal inputs, such as images or voice recordings, presents an opportunity to better mimic the versatility of human practitioners. Enabling models to process and interpret medical images, for instance, could enhance their ability to provide comprehensive answers when textual information alone is insufficient.

Integrating LLMs with medical knowledge bases and evidence-based guidelines presents another opportunity to improve accuracy. Retrieval-Augmented Generation (RAG) techniques, which combine LLMs with information retrieval systems, allow models to access up-to-date and authoritative medical information during response generation. By fetching relevant data from trusted sources like MedlinePlus or PubMed, models can provide more accurate and contextually appropriate answers, reducing the risk of misinformation.

Addressing the challenge of incomplete or ambiguous questions is crucial. Future research could explore methods for LLMs to handle incomplete information more effectively, such as by generating clarifying questions or recognizing when additional data is needed. Developing models capable of understanding implicit context—like situational factors during a pandemic—would also enhance their applicability in real-world scenarios. The study underscores the need for more comprehensive evaluation methods to assess LLM performance accurately. Future work should involve larger and more diverse sample sizes for evaluation, including complex and ambiguous questions that reflect the full spectrum of consumer inquiries. Additionally, establishing standardized benchmarks and metrics for consumer health QA can provide a clearer picture of model capabilities and areas needing improvement.

As AI technologies permeate the healthcare sector, adhering to regulatory requirements becomes essential. Future work should consider compliance with health information regulations like HIPAA in the United States or GDPR in Europe when handling sensitive data. Developing models that not only perform well but also meet legal and ethical standards will be critical for real-world deployment. Improving the interpretability of LLMs can enhance trust and facilitate their adoption in healthcare. Future research could focus on developing tools and techniques that allow users and practitioners to understand how models arrive at their conclusions. Interpretability can aid in identifying errors, biases, and areas where the model may lack sufficient knowledge.

## 5. Ethics Statement

This study is conducted solely for research purposes to evaluate the capabilities and limitations of LLMs in consumer-oriented medical question answering. While LLMs show potential for helping users find medical information, this research does not imply that these models can, or should, replace professional medical expertise. In conducting this research, a de-identified medical dataset is used to ensure privacy and data security. Any examples drawn from this dataset are paraphrased before inclusion in the paper to prevent the direct dissemination of real patient information.

## 6. Declaration of Generative AI and AI-Assisted Technologies

During the preparation of this manuscript, the authors used ChatGPT, based on the GPT-4 model, to improve the flow of the text, correct grammatical errors, and enhance the clarity of the writing. The language model was not used to generate content, citations, or verify facts. After using this tool, the authors thoroughly reviewed and edited the content to ensure accuracy, validity, and originality, and take full responsibility for the final version of the manuscript.

## 7. Acknowledgements

Gabriel Vald provided valuable assistance towards the development of this study.

## 8. Credit Author Statement

**Moaiz Abrar**: Conceptualization, Methodology, Software, Validation, Formal analysis, Investigation, Data Curation, Writing - Original Draft, and Visualization. **Yusuf Sermet**: Conceptualization, Methodology, Writing - Review & Editing, Investigation, Validation. **Ibrahim Demir**: Writing - Review & Editing, Project administration, Supervision, Funding acquisition, and Resources.